# Training a Multilingual Sportscaster:
# Using Perceptual Context to Learn Language

**David L. Chen**                                          DLCC@CS.UTEXAS.EDU
**Joohyun Kim**                                          SCIMITAR@CS.UTEXAS.EDU
**Raymond J. Mooney**                                          MOONEY@CS.UTEXAS.EDU
*Department of Computer Science*
*The University of Texas at Austin*
*1 University Station C0500, Austin TX 78712, USA*

## Abstract

We present a novel framework for learning to interpret and generate language using only perceptual context as supervision. We demonstrate its capabilities by developing a system that learns to sportscast simulated robot soccer games in both English and Korean without any language-specific prior knowledge. Training employs only ambiguous supervision consisting of a stream of descriptive textual comments and a sequence of events extracted from the simulation trace. The system simultaneously establishes correspondences between individual comments and the events that they describe while building a translation model that supports both parsing and generation. We also present a novel algorithm for learning which events are worth describing. Human evaluations of the generated commentaries indicate they are of reasonable quality and in some cases even on par with those produced by humans for our limited domain.

## 1. Introduction

Most current natural language processing (NLP) systems are built using statistical learning algorithms trained on large annotated corpora. However, annotating sentences with the requisite parse trees (Marcus, Santorini, & Marcinkiewicz, 1993), word senses (Ide & Jéronis, 1998) and semantic roles (Kingsbury, Palmer, & Marcus, 2002) is a difficult and expensive undertaking. By contrast, children acquire language through exposure to linguistic input in the context of a rich, relevant, perceptual environment. Also, by connecting words and phrases to objects and events in the world, the semantics of language is grounded in perceptual experience (Harnad, 1990). Ideally, a machine learning system would be able to acquire language in a similar manner without explicit human supervision. As a step in this direction, we present a system that can describe events in a simulated soccer game by learning only from sample language commentaries paired with traces of simulated activity without any language-specific prior knowledge. A screenshot of our system with generated commentary is shown in Figure 1.

While there has been a fair amount of research on "grounded language learning" (Roy, 2002; Bailey, Feldman, Narayanan, & Lakoff, 1997; Barnard, Duygulu, Forsyth, de Freitas, Blei, & Jordan, 2003; Yu & Ballard, 2004; Gold & Scassellati, 2007), most of the focus has been on dealing with raw perceptual data rather than language issues. Many of these systems aimed to learn meanings of words and phrases rather than interpreting entire sentences. Some more recent work has dealt with fairly complex language data (Liang, Jordan, & Klein, 2009; Branavan, Chen, Zettlemoyer, &

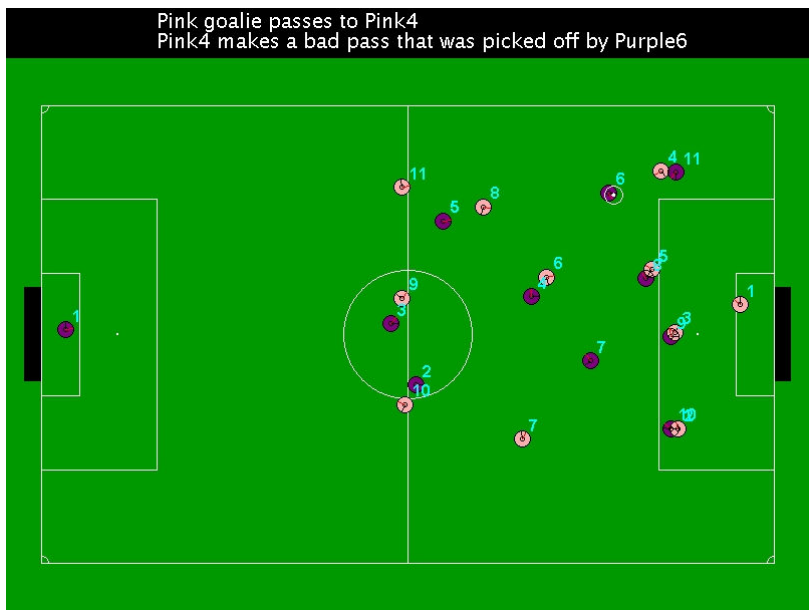

Figure 1: Screenshot of our commentator system

Barzilay, 2009) but do not address all three problems of alignment, semantic parsing, and natural language generation. In contrast, our work investigates how to build a complete language learning system using parallel data from the perceptual context. We study the problem in a simulated environment that retains many of the important properties of a dynamic world with multiple agents and actions while avoiding many of the complexities of robotics and computer vision. Specifically, we use the RoboCup simulator (Chen, Foroughi, Heintz, Kapetanakis, Kostiadis, Kummeneje, Noda, Obst, Riley, Steffens, Wang, & Yin, 2003) which provides a fairly detailed physical simulation of robot soccer. While several groups have constructed RoboCup commentator systems (André, Binsted, Tanaka-Ishii, Luke, Herzog, & Rist, 2000) that provide a textual natural-language (NL) transcript of the simulated game, their systems use manually-developed templates and are not based on learning.

Our commentator system learns to semantically interpret and generate language in the RoboCup soccer domain by observing an on-going commentary of the game paired with the evolving simulator state. By exploiting existing techniques for abstracting a symbolic description of the activity on the field from the detailed states of the physical simulator (André et al., 2000), we obtain a pairing of natural language with a symbolic description of the perceptual context in which it was uttered. However, such training data is highly ambiguous because each comment usually co-occurs with several events in the game. We integrate and enhance existing methods for learning semantic parsers and NL generators (Kate & Mooney, 2007; Wong & Mooney, 2007) in order to learn to understand and generate language from such ambiguous training data. We also develop a system that, from the same ambiguous training data, learns which events are worth describing, so that it can also perform *strategic generation*, that is, deciding *what* to say as well as how to say it (*tactical generation*). [1]

---

We evaluate our system and demonstrate its language-independence by training it to generate commentaries in both English and Korean. Experiments on test data (annotated for evaluation purposes only) demonstrate that the system learns to accurately semantically parse sentences, generate sentences, and decide which events to describe. Finally, subjective human evaluation of commentated game clips demonstrate that in our limited domain, the system generates sportscasts that are in some cases similar in quality to those produced by humans.

There are three main contributions we make in this paper. First, we explore the possibility of learning grounded language models from the perceptual context in the form of ambiguous parallel data. Second, we investigate several different methods of disambiguating this data and determined that using a combined score that includes both tactical and strategic generation scores performed the best overall. Finally, we built a complete system that learns how to sportscast in multiple languages. We carefully verified through automatic and human evaluations that the system is able to perform several tasks including disambiguating the training data, semantic parsing, tactical and strategic generation. While the language involved in this work is restricted compared to handcrafted commercial sportscasting systems, our goal was to demonstrate the feasibility of learning a grounded language system with no language-specific prior knowledge.

The remainder of the paper is structured as follows. Section 2 provides background on previous work that we utilize and extend to build our system. Section 3 describes the sportscasting data we collected to train and test our approach. Section 4 and Section 5 present the details of our basic methods for learning tactical and strategic generation, respectively, and some initial experimental results. Section 6 discusses extensions to the basic system that incorporate information from strategic generation into the process of disambiguating the training data. Section 7 presents experimental results on initializing our system with data disambiguated by a recent method for aligning language with facts to which it may refer. Section 8 discusses additions that try to detect "superfluous" sentences that do not refer to any extracted event. Section 9 presents human evaluation of our automatically generated sportscasts. Section 10 reviews related work, Section 11 discusses future work, and Section 12 presents our conclusions.

## 2. Background

Systems for learning semantic parsers induce a function that maps natural-language (NL) sentences to *meaning representations* (MRs) in some formal logical language. Existing work has focused on learning from a supervised corpus in which each sentence is manually annotated with its correct MR (Mooney, 2007; Zettlemoyer & Collins, 2007; Lu, Ng, Lee, & Zettlemoyer, 2008; Jurcicek, Gasic, Keizer, Mairesse, Thomson, & Young, 2009). Such human annotated corpora are expensive and difficult to produce, limiting the utility of this approach. Kate and Mooney (2007) introduced an extension to one such system, KRISP (Kate & Mooney, 2006), so that it can learn from ambiguous training data that requires little or no human annotation effort. However, their system was unable to *generate* language which is required for our sportscasting task. Thus, we enhanced another system called WASP (Wong & Mooney, 2006) that is capable of language generation as well as semantic parsing in a similar manner to allow it to learn from ambiguous supervision. We briefly describe the previous systems below. All of these systems assume they have access to a formal deterministic *context-free grammar* (CFG) that defines the formal *meaning representation language* (MRL). Since MRLs are formal computer-interpretable languages, such a grammar is usually easily available.

## 2.1 KRISP and KRISPER

KRISP (Kernel-based Robust Interpretation for Semantic Parsing) (Kate & Mooney, 2006) uses *support vector machines* (SVMs) with string kernels to build semantic parsers. SVMs are state-of-the-art machine learning methods that learn maximum-margin separators to prevent over-fitting in very high-dimensional data such as natural language text (Joachims, 1998). They can be extended to non-linear separators and non-vector data by exploiting *kernels* that implicitly create an even higher dimensional space in which complex data is (nearly) linearly separable (Shawe-Taylor & Cristianini, 2004). Recently, kernels over strings and trees have been effectively applied to a variety of problems in text learning and NLP (Lodhi, Saunders, Shawe-Taylor, Cristianini, & Watkins, 2002; Zelenko, Aone, & Richardella, 2003; Collins, 2002; Bunescu & Mooney, 2005). In particular, KRISP uses the string kernel introduced by Lodhi et al. (2002) to classify substrings in an NL sentence.

First, KRISP learns classifiers that recognize when a word or phrase in an NL sentence indicates that a particular concept in the MRL should be introduced into its MR. It uses production rules in the MRL grammar to represent semantic concepts, and it learns classifiers for each production that classify NL substrings as indicative of that production or not. When semantically parsing a sentence, each classifier estimates the probability of each production covering different substrings of the sentence. This information is then used to compositionally build a complete MR for the sentence. Given the partial matching provided by string kernels and the over-fitting prevention provided by SVMs, KRISP has been experimentally shown to be particularly robust to noisy training data (Kate & Mooney, 2006).

KRISPER (Kate & Mooney, 2007) is an extension to KRISP that handles ambiguous training data, in which each sentence is annotated with a *set* of potential MRs, only one of which is correct. Pseudocode for the method is shown in Algorithm 1. It employs an iterative approach analogous to *expectation maximization* (EM) (Dempster, Laird, & Rubin, 1977) that improves upon the selection of the correct NL–MR pairs in each iteration. In the first iteration (lines 3-9), it assumes that all of the MRs paired with a sentence are correct and trains KRISP with the resulting noisy supervision. In subsequent iterations (lines 11-27), KRISPER uses the currently trained parser to score each potential NL–MR pair, selects the most likely MR for each sentence, and retrains the parser on the resulting disambiguated supervised data. In this manner, KRISPER is able to learn from the type of weak supervision expected for a grounded language learner exposed only to sentences in ambiguous contexts. However, the system has previously only been tested on artificially corrupted or generated data.

## 2.2 WASP and WASP$^{-1}$

WASP (Word-Alignment-based Semantic Parsing) (Wong & Mooney, 2006) uses state-of-the-art *statistical machine translation* (SMT) techniques (Brown, Cocke, Della Pietra, Della Pietra, Jelinek, Lafferty, Mercer, & Roossin, 1990; Yamada & Knight, 2001; Chiang, 2005) to learn semantic parsers. SMT methods learn effective machine translators by training on *parallel corpora* consisting of human translations of documents into one or more alternative natural languages. The resulting translators are typically significantly more effective than manually developed systems and SMT has become the dominant approach to machine translation. Wong and Mooney (2006) adapted such methods to learn to translate from NL to MRL rather than from one NL to another.

First, an SMT *word alignment* system, GIZA++ (Och & Ney, 2003; Brown, Della Pietra, Della Pietra, & Mercer, 1993), is used to acquire a bilingual lexicon consisting of NL substrings coupled

---

**Algorithm 1** KRISPER

---

**input** sentences $S$ and their associated sets of meaning representations $MR(s)$
**output** *BestExamplesSet*, a set of NL-MR pairs,
$\quad\quad$ *SemanticModel*, a KRISP semantic parser

1:
2: **main**
3: $\quad$ //Initial training loop
4: $\quad$ **for** sentence $s_i \in S$ **do**
5: $\quad\quad$ **for** meaning representation $m_j \in MR(s_i)$ **do**
6: $\quad\quad\quad$ add $(s_i, m_j)$ to *InitialTrainingSet*
7: $\quad\quad$ **end for**
8: $\quad$ **end for**
9: $\quad$ *SemanticModel* = Train(*InitialTrainingSet*)
10:
11: $\quad$ //Iterative retraining
12: $\quad$ **repeat**
13: $\quad\quad$ **for** sentence $s_i \in S$ **do**
14: $\quad\quad\quad$ **for** meaning representation $m_j \in MR(s_i)$ **do**
15: $\quad\quad\quad\quad$ $m_j.score = Evaluate(s_i, m_j, SemanticModel)$
16: $\quad\quad\quad$ **end for**
17: $\quad\quad$ **end for**
18: $\quad\quad$ *BestExampleSet* $\leftarrow$ The set of consistent examples $T = \{(s, m) | s \in S, m \in MR(s)\}$ such that $\sum_T m.score$ is maximized
19: $\quad\quad$ *SemanticModel* = $Train(BestExamplesSet)$
20: $\quad$ **until** Convergence or MAX_ITER reached
21: **end main**
22:
23: **function** Train(*TrainingExamples*)
24: $\quad$ Train KRISP on the unambiguous *TrainingExamples*
25: $\quad$ **return** The trained KRISP semantic parser
26: **end function**
27:
28: **function** Evaluate($s$, $m$, *SemanticModel*)
29: $\quad$ Use the KRISP semantic parser *SemanticModel* to find a derivation of meaning representation $m$ from sentence $s$
30: $\quad$ **return** The parsing score
31: **end function**

---

with their translations in the target MRL. As formal languages, MRLs frequently contain many purely syntactic tokens such as parentheses or brackets, which are difficult to align with words in NL. Consequently, we found it was much more effective to align words in the NL with productions of the MRL grammar used in the parse of the corresponding MR. Therefore, GIZA++ is used to produce an N to 1 alignment between the words in the NL sentence and a sequence of MRL productions corresponding to a top-down left-most derivation of the corresponding MR.

Complete MRs are then formed by combining these NL substrings and their translations using a grammatical framework called *synchronous CFG* (SCFG) (Aho & Ullman, 1972), which forms the basis of most existing *syntax-based* SMT (Yamada & Knight, 2001; Chiang, 2005). In an SCFG, the right hand side of each production rule contains *two* strings, in our case one in NL and the other in MRL. Derivations of the SCFG simultaneously produce NL sentences and their corresponding MRs. The bilingual lexicon acquired from word alignments over the training data is used to construct a set of SCFG production rules. A probabilistic parser is then produced by training a maximum-entropy model using EM to learn parameters for each of these SCFG productions, similar to the methods used by Riezler, Prescher, Kuhn, and Johnson (2000), and Zettlemoyer and Collins (2005). To translate a novel NL sentence into its MR, a probabilistic chart parser (Stolcke, 1995) is used to find the most probable synchronous derivation that generates the given NL, and the corresponding MR generated by this derivation is returned.

Since SCFGs are symmetric, they can be used to *generate* NL from MR as well as parse NL into MR (Wong & Mooney, 2007). This allows the same learned grammar to be used for both parsing and generation, an elegant property that has important advantages (Shieber, 1988). The generation system, WASP$^{-1}$, uses a *noisy-channel model* (Brown et al., 1990):

$$\arg\max_{\mathbf{e}} \Pr(\mathbf{e}|\mathbf{f}) = \arg\max_{\mathbf{e}} \Pr(\mathbf{e})\Pr(\mathbf{f}|\mathbf{e}) \tag{1}$$

Where $\mathbf{e}$ refers to the NL string generated for a given input MR, $\mathbf{f}$. $\Pr(\mathbf{e})$ is the *language model*, and $\Pr(\mathbf{f}|\mathbf{e})$ is the *parsing model* provided by WASP's learned SCFG. The generation task is to find a sentence $\mathbf{e}$ such that (1) $\mathbf{e}$ is a good sentence a priori, and (2) its meaning is the same as the input MR. For the language model, we use a standard $n$-gram model, which is useful in ranking candidate generated sentences (Knight & Hatzivassiloglou, 1995).

## 3. Sportscasting Data

To train and test our system, we assembled human-commentated soccer games from the RoboCup simulation league (www.robocup.org). Since our focus is language learning and not computer vision, we chose to use simulated games instead of real game video to simplify the extraction of perceptual information. Based on the ROCCO RoboCup commentator's incremental event recognition module (André et al., 2000) we manually developed symbolic representations of game events and a rule-based system to automatically extract them from the simulator traces. The extracted events mainly involve actions with the ball, such as kicking and passing, but also include other game information such as whether the current playmode is kickoff, offside, or corner kick. The events are represented as atomic formulas in predicate logic with timestamps. These logical facts constitute the requisite MRs, and we manually developed a simple CFG for this formal semantic language. Details of the events detected and the complete grammar can be found in Appendix A.

For the NL portion of the data, we had humans commentate games while watching them on the simulator. We collected commentaries in both English and Korean. The English commentaries were produced by two different people while the Korean commentaries were produced by a single person. The commentators typed their comments into a text box, which were recorded with a timestamp. To construct the final ambiguous training data, we paired each comment with all of the events that occurred five seconds or less before the comment was made. Examples of the ambiguous training data are shown in Figure 2. The edges connect sentences to events to which they might refer. English translations of the Korean commentaries have been included in the figure for the reader's benefit and are not part of the actual data. Note that the use of English words for predicates and constants in the MRs is for human readability only, the system treats these as arbitrary conceptual tokens and must learn their connection to English or Korean words.

|  | English dataset | Korean dataset |
|---|---|---|
| Total # of comments | 2036 | 1999 |
| Total # of words | 11742 | 7941 |
| Vocabulary size | 454 | 344 |
| Avg. words per comment | 5.77 | 3.97 |

Table 1: Word statistics for the English and Korean datasets

|  | Number of events | Number of comments | | | Events per comment | | |
|---|---|---|---|---|---|---|---|
|  |  | Total | Have MRs | Have Correct MR | Max | Average | Std. Dev. |
| English dataset | | | | | | | |
| 2001 final | 4003 | 722 | 671 | 520 | 9 | 2.235 | 1.641 |
| 2002 final | 2223 | 514 | 458 | 376 | 10 | 2.403 | 1.653 |
| 2003 final | 2113 | 410 | 397 | 320 | 12 | 2.849 | 2.051 |
| 2004 final | 2318 | 390 | 342 | 323 | 9 | 2.729 | 1.697 |
| Korean dataset | | | | | | | |
| 2001 final | 4003 | 673 | 650 | 600 | 10 | 2.138 | 2.076 |
| 2002 final | 2223 | 454 | 444 | 419 | 12 | 2.489 | 3.083 |
| 2003 final | 2113 | 412 | 396 | 369 | 10 | 2.551 | 3.672 |
| 2004 final | 2318 | 460 | 423 | 375 | 9 | 2.601 | 2.593 |

Table 2: Alignment statistics for the English and Korean datasets. Some comments do not have correct meaning representations associated with them and are essentially noise in the training data (18% of the English dataset and 8% of the Korean dataset). Moreover, on average there are more than 2 possible events linked to each comment so over half of these links are incorrect.

We annotated a total of four games, namely, the finals for the RoboCup simulation league for each year from 2001 to 2004. Word statistics about the data are shown in Table 1. While the sentences are fairly short due to the nature of sportscasts, this data provides challenges in the form of synonyms (e.g. "Pink1", "PinkG" and "pink goalie" all refer to the same player) and polysemes (e.g. "kick" in "kicks toward the goal" refers to a kick event whereas "kicks to Pink3" refers to a pass event.) Alignment statistics for the datasets are shown in Table 2. The 2001 final has almost twice the number of events as the other games because it went into double overtime.

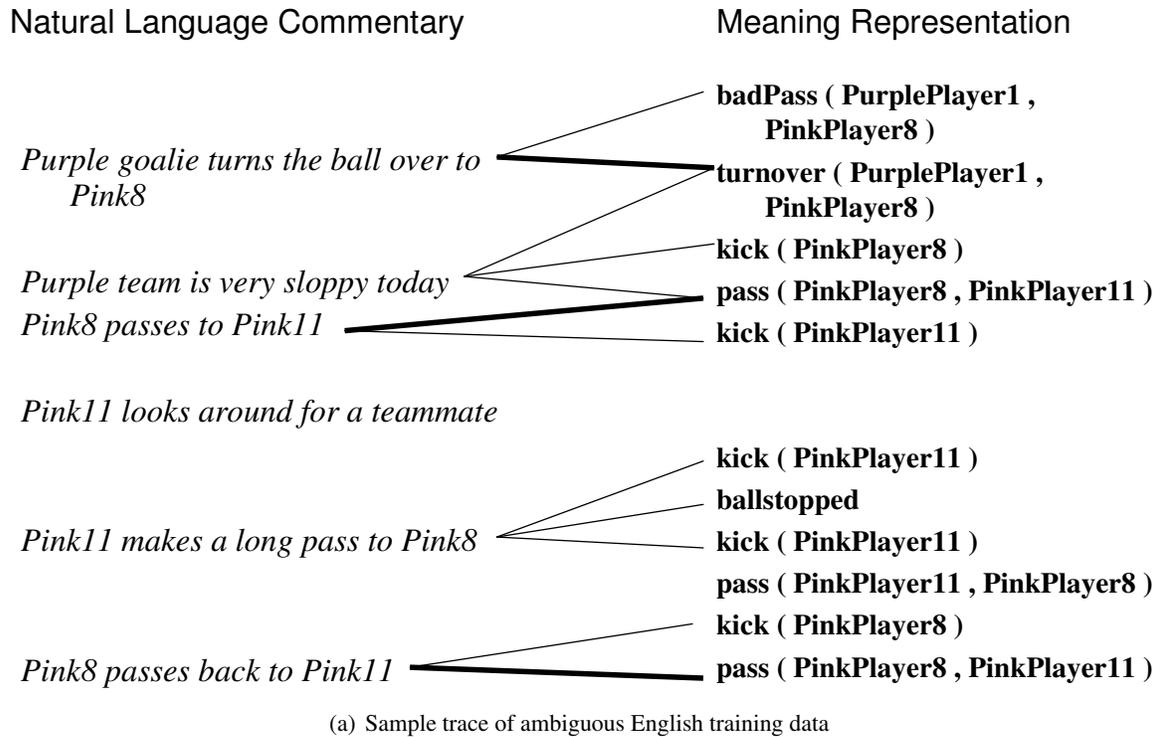

(a) Sample trace of ambiguous English training data

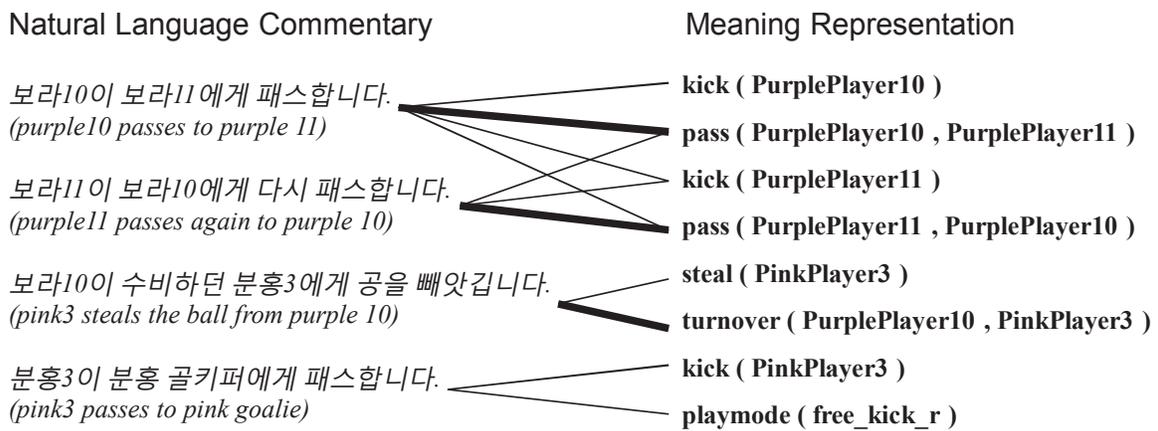

(b) Sample trace of ambiguous Korean training data

Figure 2: Examples of our training data. Each of the outgoing edges from the comments indicate a possibly associated meaning representations considered by our system. The bold links indicate correct matches between the comments and the meaning representations.

For evaluation purposes only, a gold-standard matching was produced by examining each comment manually and selecting the correct MR if it exists. The matching is only approximate because sometimes the comments contain more information than present in the MRs. For example, a comment might describe the location and the length of a pass while the MR captures only the participants of a pass. The bold lines in Figure 2 indicate the annotated correct matches in our sample data. Notice some sentences do not have correct matches (about one fifth of the English data and one tenth of the Korean data). For example, the sentence "Purple team is very sloppy today" in Figure 2(a) cannot be represented in our MRL and consequently does not have a corresponding correct MR. For another example, the Korean sentence with the translation "pink3 passes to pink goalie" in Figure 2(b) can be represented in our MRL, but does not have a correct match due to the incomplete event detection. A free kick was called while pink3 was passing to the pink goalie so the pass event was not retrieved. Finally, in the case of the sentence "Pink11 makes a long pass to Pink8" in Figure 2(a), the correct MR falls outside of the 5-second window. For each game, Table 2 shows the total number of NL sentences, the number of these that have at least one recent extracted event to which it *could* refer, and the number of these that actually *do* refer to one of these recent extracted events. The maximum, average, and standard deviation for the number of recent events paired with each comment is also given.

## 4. Learning Tactical Generation from Ambiguous Supervision

While existing systems are capable of solving parts of the sportscasting problem, none of them are able to perform the whole task. We need a system that can both deal with ambiguous supervision like KRISPER and generate language like WASP. We introduce three systems below that can do both. An overview of the differences between the existing systems and the new systems we present are shown in Table 3.

All three systems introduced here are based on extensions to WASP, our underlying language learner. The main problem we need to solve is to disambiguate the training data so that we can train WASP as before to create a language generator. Each of the new system uses a different disambiguation criteria to determine the best matching between the NL sentences and the MRs.

### 4.1 WASPER

The first system is an extension of WASP in a manner similar to how KRISP was extended to create KRISPER. It uses an EM-like retraining to handle ambiguously annotated data, resulting in a system we call WASPER. In general, any system that learns semantic parsers can be extended to handle ambiguous data as long as it can produce confidence levels for given NL–MR pairs. Given a set of sentences $s \in S$ and the set of MRs associated with each sentence $MR(s)$, we disambiguate the data by finding pairs $(s, m), s \in S$ and $m \in MR(s)$ such that $m = \arg\max_m Pr(m|s)$. Although probability is used here, a ranking of the relative potential parses would suffice. The pseudocode for WASPER is shown in Algorithm 2. The only difference compared to the KRISPER pseudocode is that we now use a WASP semantic parser instead of a KRISP parser. Also, we produce a WASP language generator as well which is the desired final output for our task.

| Algorithm | Underlying learner | Generate? | Ambiguous data? | Disambiguation criteria |
|-----------|-------------------|-----------|-----------------|------------------------|
| KRISP | SVM | No | No | n/a |
| KRISPER | KRISP | No | Yes | KRISP's parsing score |
| WASP | GIZA to align words, and MR tokens, then learn probalistic SCFG | Yes | No | n/a |
| WASPER | WASP | Yes | Yes | WASP's parsing score |
| KRISPER-WASP | First disambiguate with KRISPER, then train WASP | Yes | Yes | KRISP's parsing score |
| WASPER-GEN | WASP | Yes | Yes | NIST score of best NL given MR |

Table 3: Overview of the various learning systems presented. The first three algorithms are existing systems. We introduce the last three systems that are able to both learn from ambiguous training data and acquire a language generator. They differ in how they disambiguate the training data.

---

**Algorithm 2** WASPER

**input** sentences $S$ and their associated sets of meaning representations $MR(s)$
**output** $BestExamplesSet$, a set NL-MR pairs,
    $SemanticModel$, a WASP semantic parser/language generator
  1: **main**
  2:     same as Algorithm 1
  3: **end main**
  4:
  5: **function** Train($TrainingExamples$)
  6:     Train WASP on the unambiguous $TrainingExamples$
  7:     **return** The trained WASP semantic parser/language generator
  8: **end function**
  9:
 10: **function** Evaluate($s$, $m$, $SemanticModel$)
 11:     Use the WASP semantic parser in $SemanticModel$ to find a derivation of meaning representation $m$ from sentence $s$
 12:     **return** The parsing score
 13: **end function**

---

## 4.2 KRISPER-WASP

KRISP has been shown to be quite robust at handling noisy training data (Kate & Mooney, 2006). This is important when training on the very noisy training data used to initialize the parser in KRISPER's first iteration. However, KRISPER cannot learn a language generator, which is necessary for our sportscasting task. As a result, we create a new system called KRISPER-WASP that is both good at disambiguating the training data and capable of generation. We first use KRISPER to train on the ambiguous data and produce a disambiguated training set by using its prediction for the most likely MR for each sentence. This unambiguous training set is then used to train WASP to produce both a parser and a generator.

## 4.3 WASPER-GEN

In both KRISPER and WASPER, the criterion for selecting the best NL–MR pairs during retraining is based on maximizing the probability of parsing a sentence into a particular MR. However, since WASPER is capable of both parsing and generation, we could alternatively select the best NL–MR pairs by evaluating how likely it is to *generate* the sentence from a particular MR. Thus, we built another version of WASPER called WASPER-GEN that disambiguates the training data in order to maximize the performance of *generation* rather than parsing. The pseudocode is shown in Algorithm 3. The algorithm is the same as WASPER except for the evaluation function. It uses a generation-based score rather than a parsing-based score to select the best NL–MR pairs.

Specifically, an NL–MR pair $(s, m)$ is scored by computing the NIST score, a machine translation (MT) metric, between the sentence $s$ and the best generated sentence for $m$ (lines 9-12).[2] Formally, given a set of sentences $s \in S$ and the set of MRs associated with each sentence $MR(s)$, we disambiguate the data by finding pairs $(s, m), s \in S$ and $m \in MR(s)$ such that $m = \arg \max_m NIST(s, argmax_{s'} Pr(s'|m))$.

NIST measures the precision of a translation in terms of the proportion of $n$-grams it shares with a human translation (Doddington, 2002). It has also been used to evaluate NL generation. Another popular MT metric is BLEU score (Papineni, Roukos, Ward, & Zhu, 2002), but it is inadequate for our purpose since we are comparing one short sentence to another instead of comparing whole documents. BLEU score computes the geometric mean of the $n$-gram precision for each value of $n$, which means the score is 0 if a matching $n$-gram is not found for *every* value of $n$. In the common setting in which the maximum $n$ is 4, any two sentences that do not have a matching 4-gram would receive a BLEU score of 0. Consequently, BLEU score is unable to distinguish the quality of most of our generated sentences since they are fairly short. In contrast, NIST uses an additive score and avoids this problem.

## 4.4 Experimental Evaluation

This section presents experimental results on the RoboCup data for four systems: KRISPER, WASPER, KRISPER-WASP, and WASPER-GEN. Since we are not aware of any existing systems that could learn how to semantically parse and generate language using ambiguous supervision based on perceptual context, we constructed our own lower and upper baselines using unmodified WASP. Since

---

2. A natural way to use a generation-based score would be to use the probability of an NL given a MR ($Pr(s|m)$). However, initial experiments using this metric did not produce very good results. We also tried changing WASP to maximize the joint probability instead of just the parsing probability. However, this also did not improve the results.

---

**Algorithm 3** WASPER-GEN

---

**input** sentences $S$ and their associated sets of meaning representations $MR(s)$
**output** *BestExamplesSet*, a set of NL-MR pairs,
    *SemanticModel*, a WASP semantic parser/language generator

 1: **main**
 2:    same as Algorithm 1
 3: **end main**
 4:
 5: **function** Train(*TrainingExamples*)
 6:    same as Algorithm 2
 7: **end function**
 8:
 9: **function** Evaluate($s$, $m$, *SemanticModel*)
10:    *GeneratedSentence* ← Use the WASP language generator in *SemanticModel* to produce a
      sentence from the meaning representation $m$
11:    **return** The NIST score between *GeneratedSentence* and $s$
12: **end function**

---

WASP requires unambiguous training data, we randomly pick a meaning for each sentence from its set of potential MRs to serve as our lower baseline. We use WASP trained on the *gold matching* which consists of the correct NL–MR pairs annotated by a human as the upper baseline. This represents an upper-bound on what our systems could achieve if they disambiguated the training data perfectly.

We evaluate each system on three tasks: matching, parsing, and generation. The matching task measures how well the systems can disambiguate the training data. The parsing and generation tasks measure how well the systems can translate from NL to MR, and from MR to NL, respectively.

Since there are four games in total, we trained using all possible combinations of one to three games. For matching, we measured the performance on the training data since our goal is to disambiguate this data. For parsing and generation, we tested on the games not used for training. Results were averaged over all train/test combinations. We evaluated matching and parsing using F-measure, the harmonic mean of recall and precision. Precision is the fraction of the system's annotations that are correct. Recall is the fraction of the annotations from the gold-standard that the system correctly produces. Generation is evaluated using BLEU scores which roughly estimates how well the produced sentences match with the target sentences. We treat each game as a whole document to avoid the problem of using BLEU score for sentence-level comparisons mentioned earlier. Also, we increase the number of reference sentences for each MR by using all of the sentences in the test data corresponding to equivalent MRs, e.g. if `pass(PinkPlayer7, PinkPlayer8)` occurs multiple times in the test data, all of the sentences matched to this MR in the gold matchings are used as reference sentences for this MR.

### 4.4.1 MATCHING NL AND MR

Since handling ambiguous training data is an important aspect of grounded language learning, we first evaluate how well the various systems pick the correct NL–MR pairs. Figure 3 shows the F-measure for identifying the correct set of pairs for the various systems. All of the learning systems perform significantly better than random which has a F-measure below 0.5. In both the English and Korean data, WASPER-GEN is the best system. WASPER also equals or outperforms the previous system KRISPER as well.

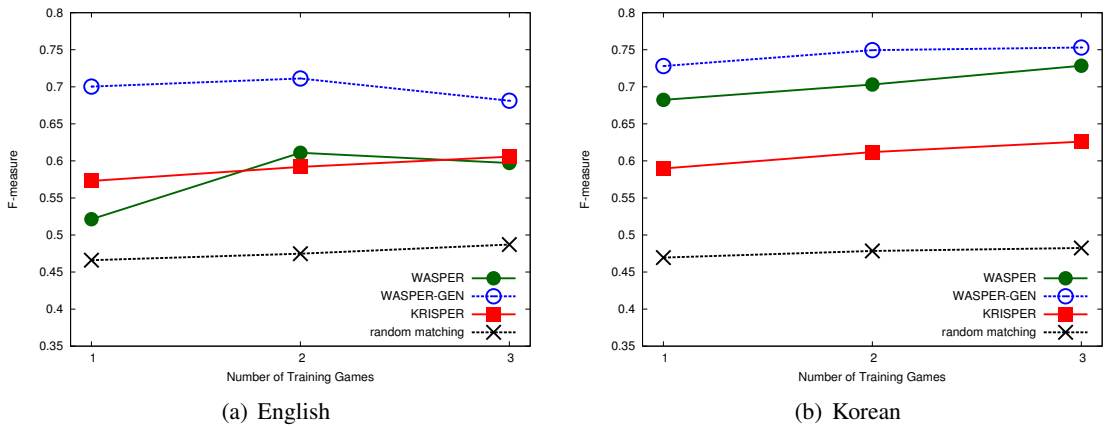

(a) English

(b) Korean

Figure 3: Matching results for our basic systems. WASPER-GEN performs the best, outperforming the existing system KRISPER on both datasets.

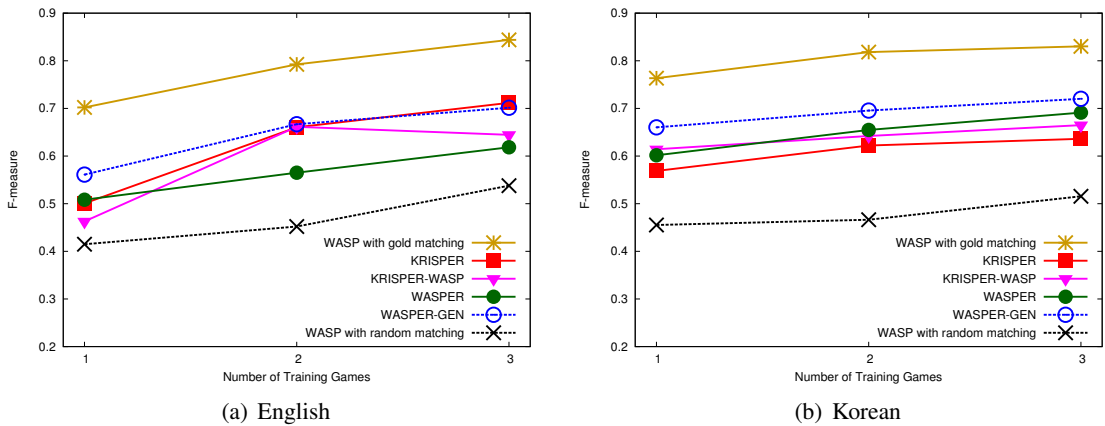

(a) English

(b) Korean

Figure 4: Semantic parsing results for our basic systems. The results largely mirrors that of the matching results with WASPER-GEN performing the best overall.

### 4.4.2 SEMANTIC PARSING

Next, we present results on the accuracy of the learned semantic parsers. Each trained system is used to parse and produce an MR for each sentence in the test set that has a correct MR in the gold-standard matching. A parse is considered correct if and only if it matches the gold standard exactly. Parsing is a fairly difficult task because there is usually more than one way to describe the same event. For example, "Player1 passes to player2" can refer to the same event as "Player1 kicks the ball to player2." Thus, accurate parsing requires learning all the different ways people describe an event. Synonymy is not limited to verbs. In our data, "Pink1", "PinkG" and "pink goalie" all refer to player1 on the pink team. Since we are not providing the systems with any prior knowledge, they have to learn all these different ways of referring to the same entity.

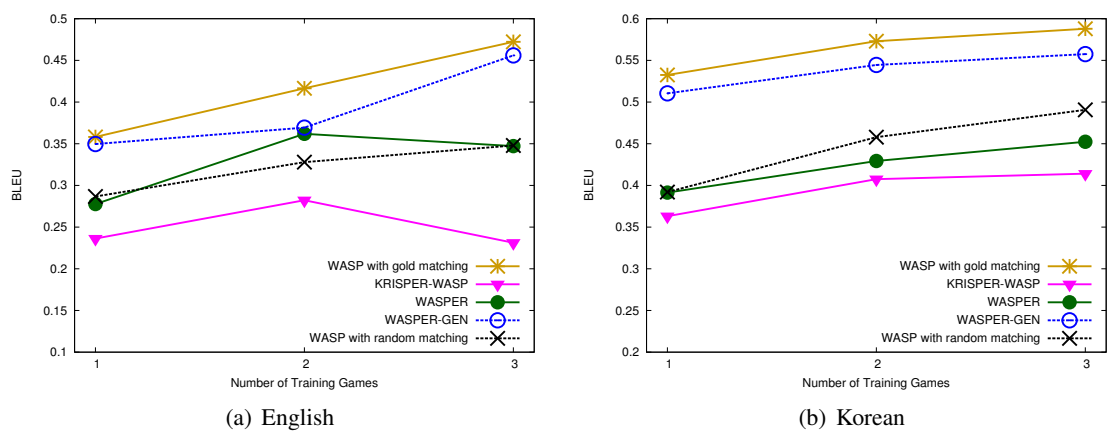

(a) English          (b) Korean

Figure 5: Tactical generation results for our basic systems. While the relative performances of the various systems change, WASPER-GEN is still the best system.

The parsing results shown in Figure 4 generally correlate well with the matching results. Systems that did better at disambiguating the training data also did better on parsing because their supervised training data is less noisy. WASPER-GEN again does the best overall on both the English and Korean data. It is interesting to note that KRISPER did relatively well on the English data compared to its matching performance. This is because KRISP is more robust to noise than WASP (Kate & Mooney, 2006) so even though it is trained on a noisier set of data than WASPER-GEN it still produced a comparable parser.

### 4.4.3 GENERATION

The third evaluation task is generation. All of the WASP-based systems are given each MR in the test set that has a gold-standard matching NL sentence and asked to generate an NL description. The quality of the generated sentence is measured by comparing it to the gold-standard using BLEU scoring.

This task is more tolerant to noise in the training data than parsing because the system only needs to learn *one way* to accurately describe an event. This property is reflected in the results, shown in Figure 5, where even the baseline system, WASP with random matching, does fairly well, outperforming KRISPER-WASP on both datasets and WASPER on the Korean data. As the number of event types is fairly small, only a relatively small number of correct matchings is required to perform this task well as long as each event type is associated with some correct sentence pattern more often than any other sentence pattern.

As with the other two tasks, WASPER-GEN is the best system on this task. One possible explanation of WASPER-GEN's superior performance stems from its disambiguation objective function. Systems like WASPER and KRISPER-WASP that use parsing scores attempt to learn a good translation model for each sentence pattern. On the other hand, WASPER-GEN only tries to learn a good translation model for each MR pattern. Thus, WASPER-GEN is more likely to converge on a good model as there are fewer MR patterns than sentence patterns. However, it can be argued that learning good translation models for each sentence pattern will help in producing more varied commentaries, a quality that is not captured by the BLEU score. Another possible advantage for WASPER-GEN is that it uses a softer scoring function. While the probabilities of parsing from a particular sentence to a MR can be sensitive to noise in the training data, WASPER-GEN only looks at the top generated sentences for each MR. Even with noise in the data, this top generated sentence remains relatively constant. Moreover, minor variations of this sentence do not change the results dramatically since the NIST score allows for partial matching.

## 5. Learning for Strategic Generation

A language generator alone is not enough to produce a sportscast. In addition to tactical generation which is deciding *how* to say something, a sportscaster must also preform *strategic generation* which is choosing *what* to say (McKeown, 1985).

We developed a novel method for learning which events to describe. For each event type (i.e. for each predicate like `pass`, or `goal`), the system uses the training data to estimate the probability that it is mentioned by the sportscaster. Given the gold-standard NL–MR matches, this probability is easy to estimate; however, the learner does not know the correct matching. Instead, the system must estimate the probabilities from the ambiguous training data. We compare two basic methods for estimating these probabilities.

The first method uses the *inferred* NL–MR matching produced by the language-learning system. The probability of commenting on each event type, $e_i$, is estimated as the percentage of events of type $e_i$ that have been matched to *some* NL sentence.

The second method, which we call Iterative Generation Strategy Learning (IGSL), uses a variant of EM, treating the matching assignments as hidden variables, initializing each match with a prior probability, and iterating to improve the probability estimates of commenting on each event type. Unlike the first method, IGSL uses information about MRs not explicitly associated with any sentence in training. Algorithm 4 shows the pseudocode. The main loop alternates between two steps:

1. Calculating the expected probability of each NL–MR matching given the current model of how likely an event is commented on (line 6)

2. Update the prior probability that an event type is mentioned by a human commentator based on the matchings (line 9).

In the first iteration, each NL–MR match is assigned a probability inversely proportional to the amount of ambiguity associated with the sentence ($\sum_{e \in Event(s)} Pr(e) = |Event(s)|$). For example, a sentence associated with five possible MRs will assign each match a probability of $\frac{1}{5}$. The prior probability of mentioning an event type is then estimated as the average probability assigned to instances of this event type. Notice this process does not always guarantee a proper probability since a MR can be associated with multiple sentences. Thus, we limit the probability to be at most one. In the subsequent iterations, the probabilities of the NL–MR matchings are updated according to these new priors. We assign each match the prior probability of its event type normalized across all the associated MRs of the NL sentence. We then update the priors for each event type as before using

---

**Algorithm 4** Iterative Generation Strategy Learning

---

**input** event types $E = \{e_1, ..., e_n\}$, the number of occurrences of each event type $TotalCount(e_i)$ in the entire game trace, sentences $S$ and the event types of their associated meaning representations $Event(s)$

**output** probabilities of commenting on each event type $Pr(e_i)$

 1: Initialize all $Pr(e_i) = 1$
 2: **repeat**
 3:    **for** event type $e_i \in E$ **do**
 4:       $MatchCount = 0$
 5:       **for** sentence $s \in S$ **do**
 6:          $ProbOfMatch = \frac{\sum_{e \in Event(s) \wedge e = e_i} Pr(e)}{\sum_{e \in Event(s)} Pr(e)}$
 7:          $MatchCount = MatchCount + ProbOfMatch$
 8:       **end for**
 9:       $Pr(e_i) = min(\frac{MatchCount}{TotalCount(e_i)}, 1)$ {Ensure proper probabilities}
10:    **end for**
11: **until** Convergence or MAX_ITER reached

---

Figure 6: An example of how our strategic generation component works. At every timestep, we stochastically select an event from all the events occurring at that moment. We then decide whether to verbalize the selected event based on IGSL's estimated probability of it being commented upon.

the new estimated probabilities for the matchings. This process is repeated until the probabilities converge or a pre-specified number of iterations has occurred.

To generate a sportscast, we use the learned probabilities to determine which events to describe. For each time step, we first determine all the events that are occurring at the time. We then select one randomly based on their normalized probabilities. To avoid being overly verbose, we do not want to make a comment every time something is happening, especially if it's an event rarely commented on. Thus, we stochastically decide whether to comment on this selected event based on its probability. An example of this process is shown in Figure 6.

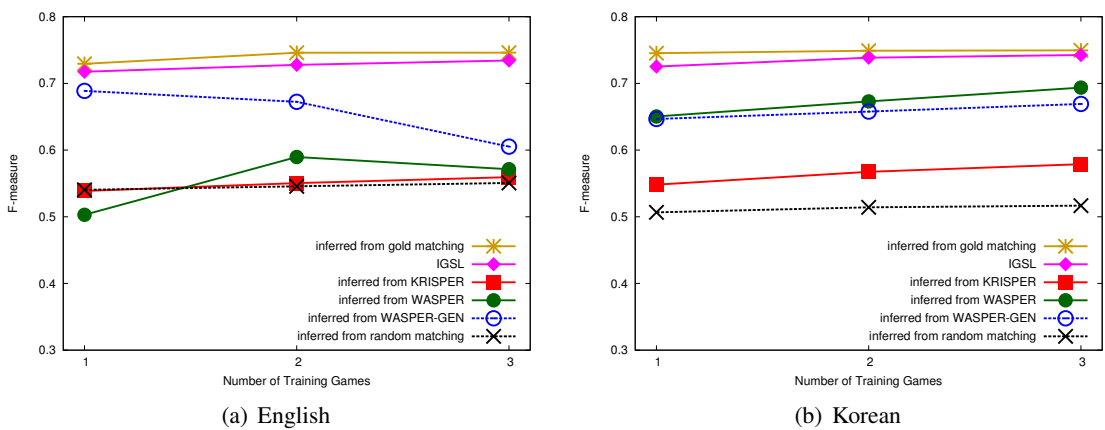

(a) English                    (b) Korean

Figure 7: Strategic generation results for our various systems. Our novel algorithm IGSL performs the best, almost on par with the upper bound which uses gold-annotated matchings.

| event | # occurrences | % commented | IGSL | inferred from WASPER-GEN |
|---|---|---|---|---|
| ballstopped | 5817 | $1.72 \times 10^{-4}$ | $1.09 \times 10^{-5}$ | 0.016 |
| kick | 2122 | 0.0 33 | 0.018 | 0.117 |
| pass | 1069 | 0.999 | 0.983 | 0.800 |
| turnover | 566 | 0.214 | 0.909 | 0.353 |
| badPass | 371 | 0.429 | 0.970 | 0.493 |

Table 4: Top 5 most frequent events, the % of times they were commented on, and the probabilities learned by the top algorithms for the English data

## 5.1 Experimental Evaluation

The different methods for learning strategic generation were evaluated based on how often the events they describe in the test data coincide with those the human decided to describe. For the first approach, results using the inferred matchings produced by Krisper, Wasper, and Wasper-Gen as well as the gold and random matching for establishing baselines are all presented in Figure 7. From the graph, it is clear that IGSL outperforms learning from the inferred matchings and actually performs at a level close to using the gold matching. However, it is important to note that we are limiting the potential of learning from the gold matching by using only the predicates to decide whether to talk about an event.

For the English data, the probabilities learned by IGSL and by inferred matchings from Wasper-Gen for the five most frequently occurring events are shown in Table 4. While Wasper-Gen learns fairly good probabilities in general, it does not do as well as IGSL for the most frequent events. This is because IGSL uses occurrences of events that are not associated with any possible comments in its training iterations. Rarely commented events such as *ballstopped* and *kick* often occur without any comments being uttered. Consequently, IGSL assigns low prior probabilities to them which lowers their chances of being matched to any sentences. On the other hand, Wasper-Gen does not use these priors and sometimes incorrectly matches comments to them. Thus, using the inferred matches from WASPER-GEN results in learning higher probabilities of commenting on these rarely commented events.

While all our methods only use the predicates of the MRs to decide whether to comment or not, they perform quite well on the data we collected. In particular, IGSL performs the best, so we use it for strategic generation in the rest of the paper.

## 6. Using Strategic Generation to Improve Matching

In this section, we explore how knowledge learned about strategic generation can be used to improve the accuracy of matching sentences to MRs. In the previous section, we described several ways to learn strategic generation, including IGSL which learns directly from the ambiguous training data. Knowing what events people tend to talk about should also help resolve ambiguities in the training data. Events that are more likely to be discussed should also be more likely to be matched to an NL sentence when disambiguating the training data. Therefore, this section describes methods that integrate strategic generation scores (such as those in Table 4) into the scoring of NL–MR pairs used in the matching process.

### 6.1 WASPER-GEN-IGSL

WASPER-GEN-IGSL is an extension of WASPER-GEN that also uses strategic generation scores from IGSL. WASPER-GEN uses NIST score to pick the best MR for a sentence by finding the MR that generates the sentence closet to the actual NL sentence. WASPER-GEN-IGSL combines tactical (NIST) and strategic (IGSL) generation scores to pick the best NL–MR pairs. It simply multiplies the NIST score and the IGSL score together to form a composite score. This new score biases the selection of matching pairs to include events that IGSL determines are, *a priori*, more likely to be discussed. This is very helpful, especially in the beginning when WASP does not produce a particularly good language generator. In many instances, the generated sentences for all of the possible MRs are equally bad and do not overlap with the target sentence. Even if generation produces a perfectly good sentence, the generation score can be unreliable because it is comparing a single sentence to a single reference that is often very short as well. Consequently, it is often difficult for WASPER-GEN to distinguish among several MRs with equal scores. On the other hand, if their event types have very different strategic generation scores, then we can default to choosing the MR with the higher prior probability of being mentioned. Algorithm 5 shows the pseudocode of WASPER-GEN-IGSL.

### 6.2 Variant of WASPER-GEN Systems

Although WASPER-GEN uses NIST score to estimate the goodness of NL–MR pairs, it could easily use any MT evaluation metric. We have already discussed the unsuitability of BLEU for comparing short individual sentences since it assigns zero to many pairs. However, NIST score also has limitations. For example, it is not normalized, which may affect the performance of WASPER-GEN-IGSL when it is combined with the IGSL score. Another limitation comes from using higher-order N-grams. Commentaries in our domain are often short, so there are frequently no higher-order N-gram matches between generated sentences and target NL sentences.

The METEOR metric (Banerjee & Lavie, 2005) was designed to resolve various weaknesses of the BLEU and NIST metrics, and it is more focused on word-to-word matches between the reference

---

**Algorithm 5** Wasper-Gen-IGSL

---

**input** sentences $S$ and their associated sets of meaning representations $MR(s)$
**output** *BestExamplesSet*, a set of NL-MR pairs,
    *SemanticModel*, a Wasp semantic parser/language generator

 1: **main**
 2:    same as Algorithm 1
 3: **end main**
 4:
 5: **function** Train(*TrainingExamples*)
 6:    same as Algorithm 2
 7: **end function**
 8:
 9: **function** Evaluate($s$, $m$, *SemanticModel*)
10:    Call Algorithm 4 to collect IGSL scores
11:    *GeneratedSentence* ← Use the Wasp language generator in *SemanticModel* to produce a
     sentence from the meaning representation $m$
12:    *TacticalGenerationScore* ← The NIST score between *GeneratedSentence* and $s$
13:    *StrategicGenerationScore* ← $Pr$(event type of $m$) from the result of Algorithm 4
14:    **return** *TacticalGenerationScore* × *StrategicGenerationScore*
15: **end function**

---

sentence and the test sentence. METEOR first evaluates uni-gram matches between the reference and the test sentence and also determines how well the words are ordered. METEOR seems more appropriate for our domain because some good generated sentences have missing adjectives or adverbs that are not critical to the meaning of the sentence but prevent higher-order N-gram matches. In addition, METEOR is normalized to always be between 0 and 1, so it may combine more effectively with IGSL scores (which are also in the range 0–1).

### 6.3 Experimental Evaluation

We evaluated the new systems, Wasper-Gen-Igsl with both NIST and METEOR scoring using the methodology from Section 4.4. The matching results are shown in Figure 8, including results for Wasper-Gen, the best system from the previous section. Both Wasper-Gen-Igsl with either NIST or METEOR scoring clearly outperforms Wasper-Gen. This indicates that strategic-generation information can help disambiguate the data. Using different MT metrics produces a less noticeable effect. There is no clear winner on the English data; however, METEOR seems to improve performance on the Korean data.

Parsing results are shown in Figure 9. As previously noted, parsing results generally mirror the matching results. Both new systems again outperform Wasper-Gen, the previously best system. And again, the English data does not show a clear advantage of using either NIST or METEOR, while the Korean data gives a slight edge to using the METEOR metric.

Results for tactical generation are shown in Figure 10. For both the English and the Korean data, the new systems come close to the performance of Wasper-Gen but do not beat it. However, the new systems do outperform Krisper-Wasp and Wasper which are not shown in the figure.

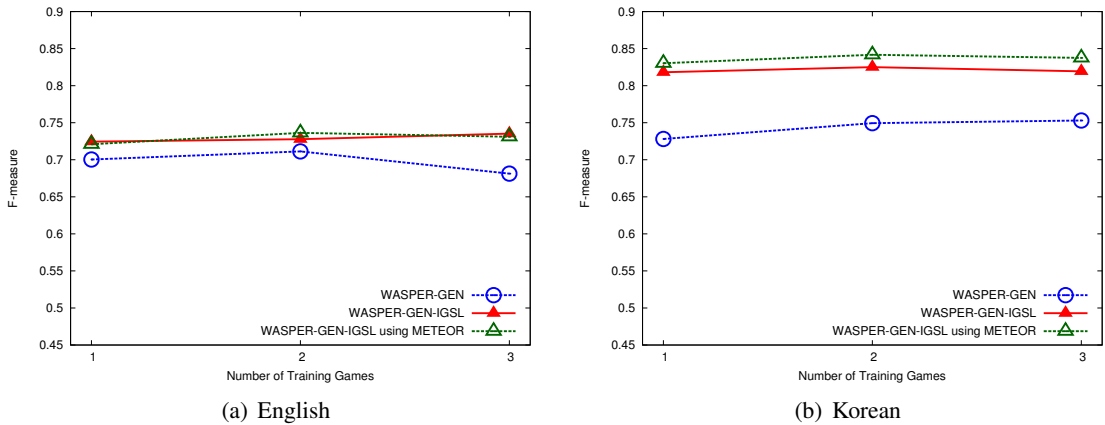

(a) English       (b) Korean

Figure 8: Matching results. Integrating strategic information improves the results over the previously best system WASPER-GEN. The choice of the MT metric used, however, makes less of an impact.

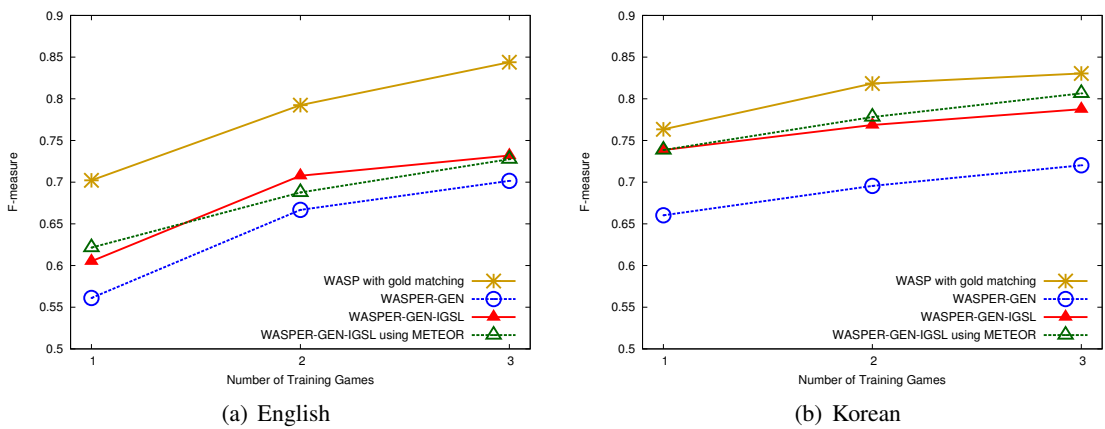

(a) English       (b) Korean

Figure 9: Semantic parsing results. The results are similar to the matching results in that integrating strategic generation information improves the performance.

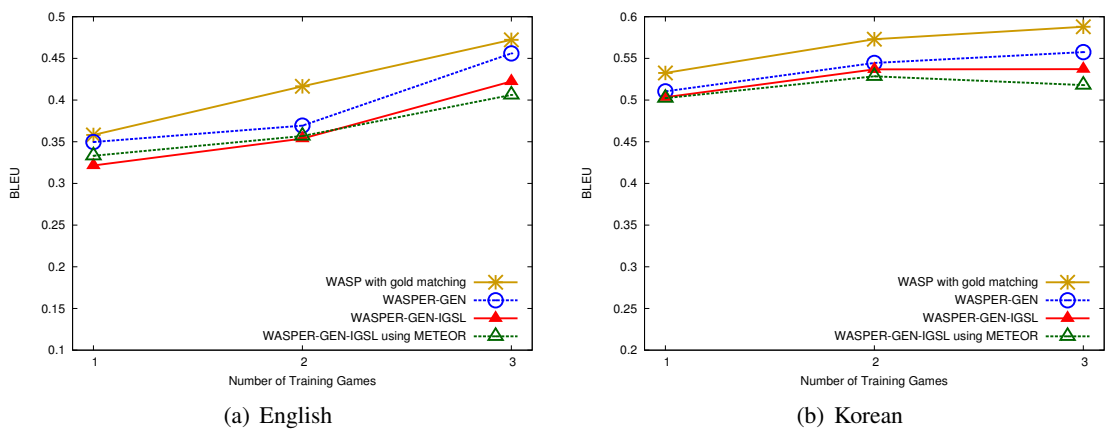

(a) English  (b) Korean

Figure 10: Tactical generation results. While the two new systems come close to the performance of WASPER-GEN, they do not beat it. However, they do outperform other systems presented earlier which are not shown in this figure.

Overall, as expected, using strategic information improves performance on the matching and semantic parsing tasks. For both the English and the Korean datasets, WASPER-GEN-IGSL and its variant using the METEOR metric clearly outperform WASPER-GEN which does not utilize strategic information. However, strategic information does not improve tactical generation. This could be due to the ceiling effect in which WASPER-GEN already performs at a level near the upper baseline. While the matching performance has improved, the generation performance has little room to grow.

## 7. Using a Generative Alignment Model

Recently, Liang et al. (2009) developed a generative model that can be used to match natural-language sentences to facts in a corresponding database to which they may refer. As one of their evaluation domains, they used our English RoboCup sportscasting data. Their method solves the matching (alignment) problem for our data, but does not address the tasks of semantic parsing or language generation. However, their generative model elegantly integrates simple strategic and tactical language generation models in order to find the overall most probable alignment of sentences and events. They demonstrated improved matching performance on our English data, generating more accurate NL–MR pairs than our best system. Thus, we were curious if their results could be used to improve our own systems, which also perform semantic parsing and generation. We also ran their code on our new Korean data but that resulted in much worse matching results compared to our best system as can be seen in Table 5.

The simplest way of utilizing their results is to use the NL–MR pairs produced by their method as supervised data for WASP. As expected, the improved NL–MR pairs for the English data resulted in improved semantic parsers as can be seen in the results in Table 6. Even for the Korean dataset, training on matchings produced by their system ended up doing fairly well even though the matching performance was poor. For tactical generation, using their matching only produced marginal improvement on the English dataset and a surprisingly large improvement on the Korean data as shown in Table 7. Overall, using the alignments produced by Liang et al.'s system resulted in good semantic parsers and tactical generators.

| Algorithm | English dataset | | Korean dataset | |
|---|---|---|---|---|
| | No initialization | Initialized | No initialization | Initialized |
| Liang et al. (2009) | 75.7 | | 69.4 | |
| WASPER | 59.7 | 79.3 | 72.8 | 76.6 |
| WASPER-GEN | 68.1 | 75.8 | 75.3 | 80.0 |
| WASPER-GEN-IGSL | 73.5 | 73.9 | 81.9 | 81.6 |
| WASPER-GEN-IGSL-METEOR | 73.1 | 75.1 | 83.8 | 84.1 |

Table 5: Matching results (F1 scores) on 4-fold cross-validation for both the English and the Korean datasets. Systems run with initialization are initialized with the matchings produced by Liang et al.'s (2009) system.

| Algorithm | English dataset | | Korean dataset | |
|---|---|---|---|---|
| | No initialization | Initialized | No initialization | Initialized |
| WASP | n/a | 80.3 | n/a | 74.01 |
| WASPER | 61.84 | 79.32 | 69.12 | 75.69 |
| WASPER-GEN | 70.15 | 77.59 | 72.02 | 77.49 |
| WASPER-GEN-IGSL | 73.19 | 73.04 | 78.75 | 75.27 |
| WASPER-GEN-IGSL-METEOR | 72.75 | 74.62 | 80.65 | 81.21 |

Table 6: Semantic parsing results (F1 scores) on 4-fold cross-validation for both the English and the Korean datasets. Systems run with initialization are initialized with the matchings produced by Liang et al.'s (2009) system.

| Algorithm | English dataset | | Korean dataset | |
|---|---|---|---|---|
| | No initialization | Initialized | No initialization | Initialized |
| WASP | n/a | 0.4580 | n/a | 0.5828 |
| WASPER | 0.3471 | 0.4599 | 0.4524 | 0.6118 |
| WASPER-GEN | 0.4560 | 0.4414 | 0.5575 | 0.6796 |
| WASPER-GEN-IGSL | 0.4223 | 0.4585 | 0.5371 | 0.6710 |
| WASPER-GEN-IGSL-METEOR | 0.4062 | 0.4353 | 0.5180 | 0.6591 |

Table 7: Tactical generation results (BLEU score) on 4-fold cross-validation for both the English and the Korean datasets. Systems run with initialization are initialized with the matchings produced by Liang et al.'s (2009) system.

In addition to training WASP with their alignment, we can also utilize their output as a better *starting point* for our own systems. Instead of initializing our iterative alignment methods with a model trained on *all* of the ambiguous NL–MR pairs, they can be initialized with the disambiguated NL–MR pairs produced by Liang et al.'s system.

Initializing the systems in this manner almost always improved the performance on all three tasks (Tables 5, 6, and 7). Moreover, the results from the best systems exceed that of simply training WASP with the alignment in all cases except for semantic parsing on the English data. Thus, combining Liang et al.'s alignment with our disambiguation techniques seems to produce the best overall results. For the English data, WASPER with initialization performs the best on both matching and generation. It does slightly worse on the semantic parsing task compared to WASP trained on Liang et al.'s alignment. For the Korean data, all the systems do better than just training WASP on the alignment. WASPER-GEN-IGSL-METEOR with initialization performs the best on matching and semantic parsing while WASPER-GEN with initialization performs the best on generation.

Overall, initializing our systems with the alignment output of Liang et al.'s generative model improved performance as expected. Starting with a cleaner set of data led to better initial semantic parsers and language generators which led to better end results. Furthermore, by incorporating a semantic parser and a tactical generator, we were able to improve on the Liang et al.'s alignments and achieve even better results in most cases.

## 8. Removing Superfluous Comments

So far, we have only discussed how to handle ambiguity in which there are multiple possible MRs for each NL sentence. During training, all our methods assume that each NL sentence matches exactly one of the potential MRs. However, some comments are *superfluous*, in the sense that they do not refer to *any* currently extracted event represented in the set of potential MRs. As previously shown in Tables 2, about one fifth of the English sentences and one tenth of the Korean sentences are superfluous in this sense.

There are many reasons for superfluous sentences. They occur naturally in language because people do not always talk about the current environment. In our domain, sportscasters often mention past events or more general information about particular teams or players. Moreover, depending on the application, the chosen MRL may not represent all of the things people talk about. For example, our RoboCup MRL cannot represent information about players who are not actively engaged with the ball. Finally, even if a sentence can be represented in the chosen MRL, errors in the perceptual system or an incorrect estimation of when an event occurred can also lead to superfluous sentences. Such perceptual errors can be alleviated to some degree by increasing the size of the window used to capture potential MRs (the previous 5 seconds in our experiments). However, this comes at the cost of increased ambiguity because it associates more MRs with each sentence.

To deal with the problem of superfluous sentences, we can eliminate the lowest-scoring NL–MR pairs (e.g. lowest parsing scores for WASPER or lowest NIST scores for WASPER-GEN). However, in order to set the pruning threshold, we need to automatically estimate the *amount* of superfluous commentary in the absence of supervised data. Notice that while this problem looks similar to the strategic generation problem (estimating how likely an MR participates in a correct matching as opposed to how likely an NL sentence participates in a correct matching), the approaches used there cannot be applied. First, we cannot use the matches inferred by the existing systems to estimate the fraction of superfluous comments since the current systems match every sentence to some MR. It is also difficult to develop an algorithm similar to IGSL due to the imbalance between NL sentences and MRs. Since there are many more MRs, there are more examples of events occurring without commentaries than vice versa.

## 8.1 Estimating the Superfluous Rate Using Internal Cross Validation

We propose using a form of internal (i.e. within the training set) cross validation to estimate the rate of superfluous comments. While this algorithm can be used in conjunction with any of our systems, we chose to implement it for KRISPER which trains much faster than our other systems. This makes it more tractable to train many different semantic parsers and choose the best one. The basic idea is to use part of the ambiguous training data to estimate the accuracy of a semantic parser even though we do not know the correct matchings. Assuming a reasonable superfluous sentence rate, we know that most of the time the correct MR is contained in the set of MRs associated with an NL sentence. Thus, we assume that a semantic parser that parses an NL sentence into one of the MRs associated with it is better than one that parses it into an MR not in the set. With this approach to estimating accuracy, we can evaluate semantic parsers learned using various pruning thresholds and pick the best one. The algorithm is briefly summarized in the following steps:

1. Split the training set into an *internal training set* and an *internal validation set*.

2. Train KRISPER $N$ times on the *internal training set* using $N$ different threshold values (eliminating the lowest scoring NL–MR pairs below the threshold in each retraining iteration in Algorithm 1).

3. Test the $N$ semantic parsers on the *internal validation set* and determine which parser is able to parse the largest number of sentences into one of their potential MRs.

4. Use the threshold value that produced the best parser in the previous step to train a final parser on the complete original training set.

## 8.2 Experiments

We evaluated the effect of removing superfluous sentences on all three tasks: matching, parsing, and generation. We present results for both KRISPER and KRISPER-WASP. For matching, we only show results for KRISPER because it is responsible for disambiguating the training data for both systems (so KRISPER-WASP's results are the same). For generation, we only show results for KRISPER-WASP, since KRISPER cannot perform generation.

The matching results shown in Figure 11 demonstrate that removing superfluous sentences does improve the performance for both English and Korean, although the difference is small in absolute terms. The parsing results shown in Figure 12 indicate that removing superfluous sentences usually improves the accuracy of both KRISPER and KRISPER-WASP marginally. As we have observed many times, the parsing results are consistent with the matching results. Finally, the tactical generation results shown in Figure 13 suggest that removing superfluous comments actually decreases performance somewhat. Once again, a potential explanation is that generation is less sensitive to noisy training data. While removing superfluous comments improves the purity of the training data, it also removes potentially useful examples. Consequently, the system does not learn how to generate sentences that were removed from the data. Overall, for generation, the advantage of having cleaner disambiguated training data is apparently outweighed by the loss of data.

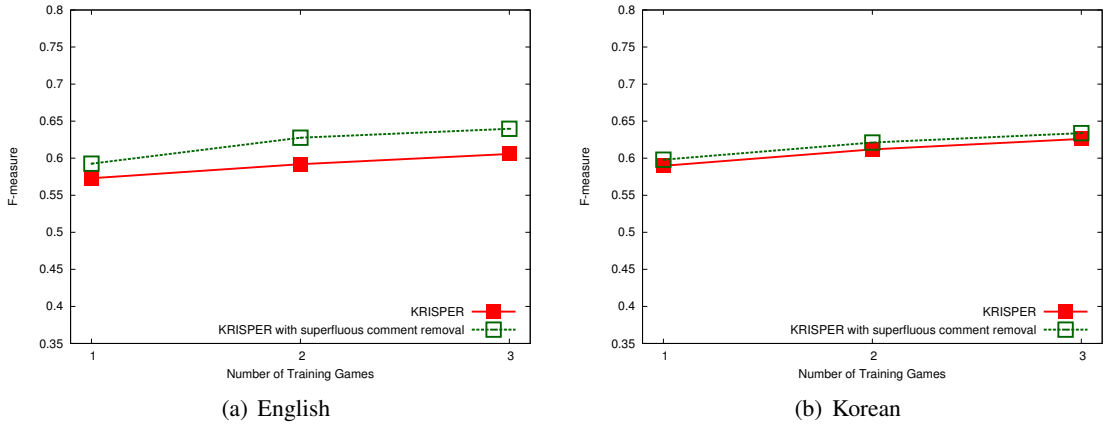

Figure 11: Matching results comparing the effects of removing superfluous comments.

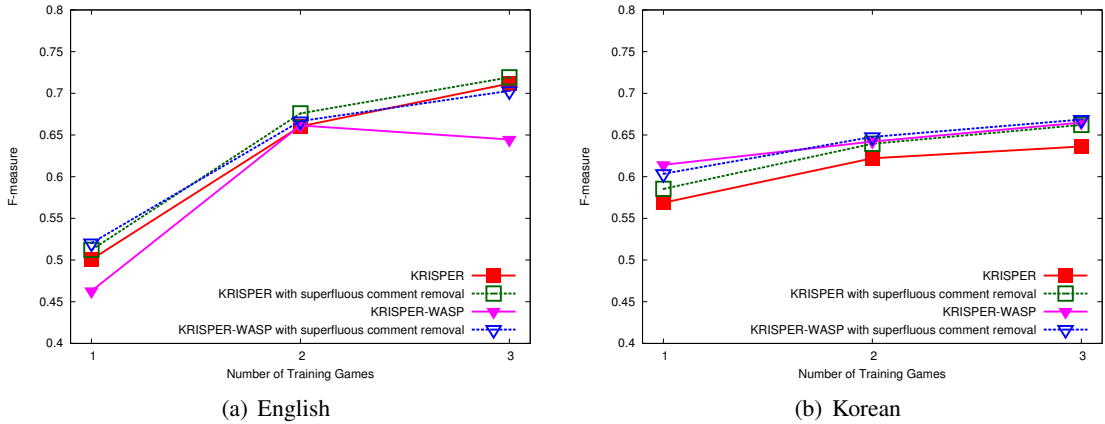

Figure 12: Semantic parsing results are improved marginally after superfluous comment removal.

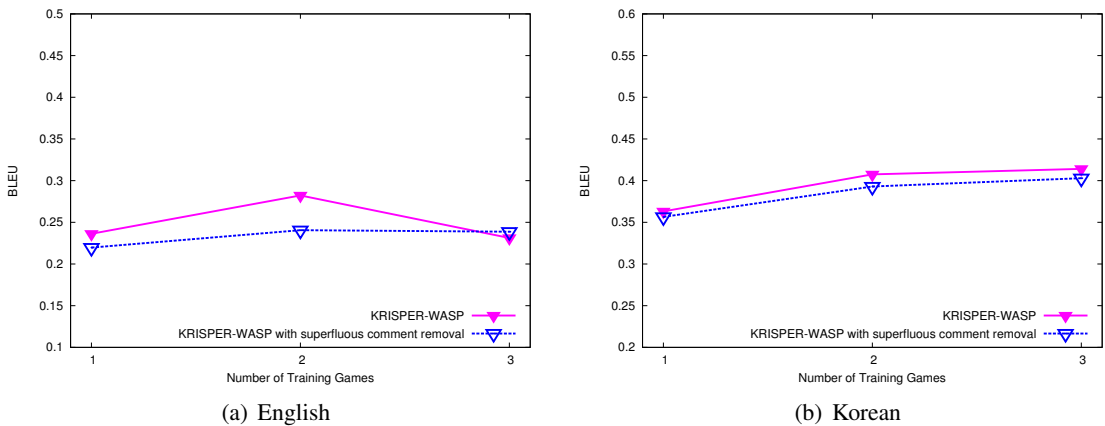

Figure 13: Tactical generation performance decreases after removing superfluous comments.

## 9. Human Subjective Evaluation

At best, automatic evaluation of generation is an imperfect approximation of human assessment. Moreover, automatically evaluating the quality of an entire generated sportscast is even more difficult. Consequently, we used Amazon's Mechanical Turk to collect human judgements of the produced sportscasts. Each human judge was shown three clips of simulated game video in one sitting. There were 8 video clips total. The 8 clips use 4 game segments of 4 minutes each, one from each of the four games (2001-2004 RoboCup finals). Each of the 4 game segments is commentated once by a human and once by our system. We use IGSL to determine the events to comment on and WASPER-GEN (our best performing system for tactical generation) to produce the commentaries. To make the commentaries more varied, we took the top 5 outputs from WASPER-GEN and chose one stochastically weighted by their scores. The system was always trained on three games, leaving out the game from which the test segment was extracted. The video clips were accompanied by commentaries that both appear as subtitles on the screen as well as audio produced by a auto-mated text to speech system [3] The videos are shown in random counter-balanced order to ensure no consistent bias toward segments being shown earlier or later. We asked the judges to score the commentaries using the following metrics:

| Score | Fluency | Semantic Correctness | Sportscasting Ability |
|-------|---------|---------------------|----------------------|
| 5 | Flawless | Always | Excellent |
| 4 | Good | Usually | Good |
| 3 | Non-native | Sometimes | Average |
| 2 | Disfluent | Rarely | Bad |
| 1 | Gibberish | Never | Terrible |

Fluency and semantic correctness, or adequacy, are standard metrics in human evaluations of NL translations and generations. Fluency measures how well the commentaries are structured, including syntax and grammar. Semantic correctness indicates whether the commentaries accurately describe what is happening in the game. Finally, sportscasting ability measures the overall quality of the sportscast. This includes whether the sportscasts are interesting and flow well. In addition to these metrics, we also asked them whether they thought the sportscast was composed by a human or a computer (*Human?*).

Since Mechanical Turk recruits judges over the Internet, we had to make sure that the judges were not assigning ratings randomly. Thus, in addition to asking them to rate each video, we also asked them to count the number of goals in each video. Incorrect responses to this question caused their ratings to be discarded. This is to ensure that the judges faithfully watched the entire clip before assigning ratings. After such pruning, there was on average 36 ratings (from 40 original ratings) for each of the 8 videos for the English data. Since it was more difficult to recruit Korean judges over the Internet, we recruited them in person and collected 7 ratings on average for each video in the Korean data. Table 8 and 9 show the results for the English and Korean data, respectively. Statistically significant results are shown in boldface.

Results are surprisingly good for the English data across all categories with the machine actually scoring higher than the human on average. However, the differences are not statistically significant based on an unpaired t-test ($p > 0.05$). Nevertheless, it is encouraging to see the machine being rated so highly. There is some variance in the human's performance since there were two different commentators. Most notably, compared to the machine, the human's performance on the 2002 final is quite good because his commentary included many details such as the position of the players, the types of passes, and comments about the overall flow of the game. On the other hand, the human's performance on the 2003 final is quite bad because the human commentator was very "mechanical" and used the same sentence pattern repeatedly. The machine performance was more even throughout although sometimes it gets lucky. For example, the machine serendipitously said "This is the beginning of an exciting match." near the start of the 2004 final clip simply because this statement was incorrectly learned to correspond to an extracted MR that is actually unrelated.

---

3. Sample video clips with sound are available on the web at `http://www.cs.utexas.edu/users/ml/clamp/sportscasting/`.

| | Commentator | Fluency | Semantic Correctness | Sportscasting Ability | Human? |
|---|---|---|---|---|---|
| 2001 final | Human | 3.74 | 3.59 | 3.15 | 20.59% |
| | Machine | 3.89 | 3.81 | **3.61** | 40.00% |
| 2002 final | Human | 4.13 | **4.58** | **4.03** | **42.11%** |
| | Machine | 3.97 | 3.74 | 3.29 | 11.76% |
| 2003 final | Human | 3.54 | 3.73 | 2.61 | 13.51% |
| | Machine | **3.89** | **4.26** | **3.37** | 19.30% |
| 2004 final | Human | 4.03 | 4.17 | 3.54 | 20.00% |
| | Machine | 4.13 | 4.38 | 4.00 | **56.25%** |
| Average | Human | 3.86 | 4.03 | 3.34 | 24.31% |
| | Machine | 3.94 | 4.03 | 3.48 | 26.76% |

Table 8: Human evaluation of overall sportscasts for English data. Bold numbers indicate statistical significance.

| | Commentator | Fluency | Semantic Correctness | Sportscasting Ability | Human? |
|---|---|---|---|---|---|
| 2001 final | Human | 3.75 | 4.13 | **4.00** | 50.00% |
| | Machine | 3.50 | 3.67 | 2.83 | 33.33% |
| 2002 final | Human | 4.17 | **4.33** | 3.83 | 83.33% |
| | Machine | 3.25 | 3.38 | 3.13 | 50.00% |
| 2003 final | Human | **3.86** | **4.29** | **4.00** | **85.71%** |
| | Machine | 2.38 | 3.25 | 2.88 | 25.00% |
| 2004 final | Human | 3.00 | 3.75 | 3.25 | 37.50% |
| | Machine | 2.71 | 3.43 | 3.00 | 14.29% |
| Average | Human | **3.66** | **4.10** | **3.76** | **62.07%** |
| | Machine | 2.93 | 3.41 | 2.97 | 31.03% |

Table 9: Human evaluation of overall sportscasts for Korean data. Bold numbers indicate statistical significance.

The results for Korean are not as impressive. The human beats the machine on average for all categories. However, the largest difference between the scores in any category is only 0.8. Moreover, the absolute scores indicate that the generated Korean sportscast is at least of acceptable quality. The judges even mistakenly thought they were produced by humans one third of the time. Part of the reason for the worse performance compared to the English data is that the Korean commentaries were fairly detailed and included events that were not extracted by our limited perceptual system. Thus, the machine simply had no way of competing because it is limited to only expressing information that is present in the extracted MRs.

We also elicited comments from the human judges to get a more qualitative evaluation. Overall, the judges thought the generated commentaries were good and accurately described the actions on the field. Picking from the top 5 generated sentences also added variability to the machine-generated sportscasts that improved the results compared with earlier experiments presented by Chen and Mooney (2008). However, the machine still sometimes misses significant plays such as scoring or corner kicks. This is because these plays happen much less frequently and often coincide with many other events (e.g. shooting the ball and kickoffs co-occur with scoring). Thus, the machine has a harder time learning about these infrequent events. Another issue concerns our representation. Many people complain about long gaps in the sportscasts or lack of details. Our event detector only concentrates on ball possession and not on positions or elapsed time. Thus, a player holding onto a ball or dribbling for a long time does not produce any events detected by our simulated perceptual system. Also, a short pass in the backfield is treated exactly the same as a long pass across the field to near the goal. Finally, people desired more colorful commentary (background information, statistics, or analysis of the game) to fill in the voids. This is a somewhat orthogonal issue since our goal was to build a play-by-play commentator that described events that were currently happening.

## 10. Related Work

In this section we review some of the related work in semantic parsing, natural language generation as well as grounded language learning.

### 10.1 Semantic Parsing

As mentioned in Section 2, existing work on semantic parser learners has focused on supervised learning where each sentence is annotated with its semantic meaning. Some semantic-parser learners additionally require either syntactic annotations (Ge & Mooney, 2005) or prior syntactic knowledge of the target language (Ge & Mooney, 2009; Zettlemoyer & Collins, 2005, 2007). Since the world never provides any direct feedback on syntactic structure, language-learning methods that require syntactic annotation are not directly applicable to grounded language learning. Therefore, methods that learn from *only* semantic annotation are critical to learning language from perceptual context.

While we use logic formulas as our MRs, the particular MRL we use contains only atomic formulas and can be equivalently represented as frames and slots. There are systems that use transformation-based learning (Jurcicek et al., 2009), or Markov logic (Meza-Ruiz, Riedel, & Lemon, 2008) to learn semantic parsers using frames and slots. In principle, our framework can be used with any semantic parser learner as long as it provides confidence scores for its parse results.

## 10.2 Natural Language Generation

There are several existing systems that sportscast RoboCup games (André et al., 2000). Given game states provided by the RoboCup simulator, they extract game events and generate real-time commentaries. They consider many practical issues such as timeliness, coherence, variability, and emotion that are needed to produce good sportscasts. However, these systems are hand-built and generate language using pre-determined templates and rules. In contrast, we concentrate on the learning problem and induce the generation components from ambiguous training data. Nevertheless, augmenting our system with some of the other components in these systems could improve the final sportscasts produced.

There is also prior work on learning a lexicon of elementary semantic expressions and their corresponding natural language realizations (Barzilay & Lee, 2002). This work uses multiple-sequence alignment on datasets that supply several verbalizations of the corresponding semantics to extract a dictionary.

Duboue and McKeown (2003) were the first to propose an algorithm for learning strategic generation automatically from data. Using semantics and associated texts, their system learns a classifier that determines whether a particular piece of information should be included for presentation or not.

There has been some recent work on learning strategic generation using reinforcement learning (Zaragoza & Li, 2005). This work involves a game setting where the speaker must aid the listener in reaching a given destination while avoiding obstacles. The game is played repeatedly to find an optimal strategy that conveys the most pertinent information while minimizing the number of messages. We consider a different problem setting where such reinforcements are not available to our strategic generation learner.

In addition, there has also been work on performing strategic generation as a collective task (Barzilay & Lapata, 2005). By considering all strategic generation decisions jointly, it captures dependencies between utterances. This creates more consistent overall output and is more consistent with how humans perform this task. Such an approach could potentially help our system produce better overall sportscasts.

## 10.3 Grounded Language Learning

One of the most ambitious end-to-end visually-grounded scene-description system is VITRA (Herzog & Wazinski, 1994) which comments on traffic scenes and soccer matches. The system first transforms raw visual data into geometrical representations. Next, a set of rules extract spatial relations and interesting motion events from those representations. Presumed intentions, plans, and plan interactions between the agents are also extracted based on domain-specific knowledge. However, since their system is hand-coded it cannot be adapted easily to new domains.

Srihari and Burhans (1994) used captions accompanying photos to help identify people and objects. They introduced the idea of visual semantics, a theory of extracting visual information and constraints from accompanying text. For example, by using caption information, the system can determine the spatial relationship between the entities mentioned, the likely size and shape of the object of interest, and whether the entity is natural or artificial. However, their system is also based on hand-coded knowledge.

Siskind (1996) performed some of the earliest work on *learning* grounded word meanings. His learning algorithm addresses the problem of ambiguous training or "referential uncertainty" for semantic lexical acquisition, but does not address the larger problems of learning complete semantic parsers and language generators.

Several robotics and computer vision researchers have worked on inferring grounded meanings of individual words or short referring expressions from visual perceptual context (e.g., Roy, 2002; Bailey et al., 1997; Barnard et al., 2003; Yu & Ballard, 2004). However, the complexity of the natural language used in this existing work is very restrictive, many of the systems use pre-coded knowledge of the language, and almost all use static images to learn language describing objects and their relations, and cannot learn language describing actions. The most sophisticated grammatical formalism used to learn syntax in this work is a finite-state hidden-Markov model. By contrast, our work exploits the latest techniques in statistical context-free grammars and syntax-based statistical machine translation that handle more of the complexities of natural language.

More recently, Gold and Scassellati (2007) built a system called TWIG that uses existing language knowledge to help it learn the meaning of new words. The robot uses partial parses to focus its attention on possible meanings of new words. By playing a game of catch, the robot was able to learn the meaning of "you" and "me" as well as "am" and "are" as identity relations.

There has also been a variety of work on learning from captions that accompany pictures or videos (Satoh, Nakamura, & Kanade, 1997; Berg, Berg, Edwards, & Forsyth, 2004). This area is of particular interest given the large amount of captioned images and video available on the web and television. Satoh et al. (1997) built a system to detect faces in newscasts. However, they use fairly simple manually-written rules to determine the entity in the picture to which the language refers. Berg et al. (2004) used a more elaborate learning method to cluster faces with names. Using the data, they estimate the likelihood of an entity appearing in a picture given its context.

Some recent work on video retrieval has focused on learning to recognize events in sports videos and connecting them to English words appearing in accompanying closed captions (Fleischman & Roy, 2007; Gupta & Mooney, 2009). However, this work only learns the connection between individual words and video events and does not learn to describe events using full grammatical sentences. To avoid difficult problems in computer vision, our work uses a simulated world where perception of complex events and their participants is much simpler.

In addition to observing events passively, there has also been work on grounded language learning in more interactive environments such as in computer video games (Gorniak & Roy, 2005). In this work, players cooperate and communicate with each in order to accomplish a certain task. The system learns to map spoken instructions to specific actions; however, it relies on existing statistical parsers and does not learn the syntax and semantics of the language from the perceptual environment alone. Kerr, Cohen, and Chang (2008) developed a system that learns grounded word-meanings for nouns, adjectives, and spatial prepositions while a human is instructing it to perform tasks in a vir-tual world; however, the system assumes an existing syntactic parser and prior knowledge of verb semantics and is unable to learn these from experience.

Recently, there has been some interest in learning how to interpret English instructions describing how to use a particular website or perform other computer tasks (Branavan et al., 2009; Lau, Drews, & Nichols, 2009). These systems learn to predict the correct computer action (pressing a button, choosing a menu item, typing into a text field, etc.) corresponding to each step in the instructions. Instead of using parallel training data from the perceptual context, these systems utilize direct matches between words in the natural language instructions and English words explicitly occurring in the menu items and computer instructions in order to establish a connection between the language and the environment.

One of the core subproblems our work addresses is matching sentences to facts in the world to which they refer. Some recent projects attempt to align text from English summaries of American football games with database records that contain statistics and events about the game (Snyder & Barzilay, 2007; Liang et al., 2009). However, Snyder and Barzilay (2007) use a supervised approach that requires annotating the correct correspondences between the text and the semantic representations. On the other hand, Liang et al. (2009) have developed an unsupervised approach using a generative model to solve the alignment problem. They also demonstrated improved results on matching sentences and events on our RoboCup English sportscasting data. However, their work does not address semantic parsing or language generation. Section 7 presents results showing how our methods can improve the NL–MR matches produced by this approach as well as use them to learn parsers and generators.

## 11. Future Work

As previously discussed, some of the limitations of the current system are due to inadequacies in the perception of events extracted from the RoboCup simulator. Some of the language commentary, particularly in the Korean data, refers to information about events that is not currently represented in the extracted MRs. For example, a player dribbling the ball is not captured by our perceptual system. The event extractor could be extended to include such information in the output representations.

Commentaries are not always about the immediate actions happening on the field. They can also refer to statistics about the game, background information, or analysis of the game. While some of these are difficult to obtain, it would be simple to augment the potential MRs to include events such as the current score or the number of turnovers, etc. While these may be difficult to learn correctly, they potentially would make the commentaries much more natural and engaging.

Some statements in the commentaries specifically refer to a pattern of activity across several recent events rather than to a single event. For example, in one of the English commentaries, the statement "Purple team is very sloppy today." appears after a series of turn-overs to the other team. The simulated perception could be extended to extract patterns of activity such as "sloppiness;" however this assumes that such concepts are predefined, and extracting many such higher-level predicates would greatly increase ambiguity in the training data. The current system assumes it already has concepts for the words it needs to learn and can perceive these concepts and represent them in MRs. However, it would be interesting to include a more "Whorfian" style of language learning (Whorf, 1964) in which an unknown word such as "sloppiness" could actually cause the creation of a new concept. For content words that do not seem to consistently correlate with any perceived event, the system could collect examples of recent activity where the word is used and try to learn a new higher-level concept that captures a regularity in these situations. For example, given examples of situations referred to as "sloppy," an inductive logic programming system (Lavrač & Džeroski, 1994) should be able to detect the pattern of several recent turnovers.

Another shortcoming of the current system is that each MR is treated independently. This fails to exploit the fact that many MRs are related to each other. For example, a pass is preceded by a kick, and a bad pass is followed by a turnover. A more natural way is to use a graphical representation to represent not only the entities and events but also the relationships between them.

Currently tactical and strategic generation in our system are only loosely coupled. However, conceptually they are much more closely related, and solving one problem should help solve the other. Initializing our system with the output from Liang et al. (2009), which uses a generative model that includes both strategic and tactical components, produced somewhat better results. However, the interaction between all these components is very loose and a tighter integration of the different pieces could yield stronger results in all the tasks.

An obvious extension to the current work is to apply it to *real* RoboCup games rather than simulated ones. Recent work by Rozinat, Zickler, Veloso, van der Aalst, and McMillen (2008) analyzes games in the RoboCup Small Size League using video from the overhead camera. By using the symbolic event trace extracted by such a real perceptual system, our methods could be applied to real-world games. Using speech recognition to accept spoken language input is another obvious extension.

We are currently exploring extending our approach to learn to interpret and generate NL instructions for navigating in a virtual environment. The system will observe one person giving English navigation instructions (e.g. "Go down the hall and turn left after you pass the chair.") to another person who follows these directions to get to a chosen destination. By collecting examples of sentences paired with the actions that were executed together with information about the local environment, the system will construct an ambiguous supervised dataset for language learning. Such an approach could eventually lead to virtual agents in games and educational simulations that can automatically learn to interpret and generate natural language instructions.

## 12. Conclusion

We have presented an end-to-end system that learns to generate natural-language sportscasts for simulated RoboCup soccer games by training on sample human commentaries paired with automatically extracted game events. By learning to semantically interpret and generate language without explicitly annotated training data, we have demonstrated that a system can learn language by simply observing linguistic descriptions of ongoing events. We also demonstrated the system's language independence by successfully training it to produce sportscasts in both English and Korean.

Dealing with the ambiguous supervision inherent in the training environment is a critical issue in learning language from perceptual context. We have evaluated various methods for disambiguating the training data in order to learn semantic parsers and language generators. Using a generation evaluation metric as the criterion for selecting the best NL–MR pairs produced better results than using semantic parsing scores when the initial training data is very noisy. Our system also learns a model for strategic generation from the ambiguous training data by estimating the probability that each event type evokes human commentary. Moreover, using strategic generation information to help disambiguate the training data was shown to improve the results. We also demonstrated that our system can be initialized with alignments produced by a different system to achieve better results than either system alone. Finally, experimental evaluation verified that the overall system learns to accurately parse and generate comments and to generate sportscasts that are competitive with those produced by humans.

## Acknowledgments

We thank Adam Bossy for his work on simulating perception for the RoboCup games. We also thank Percy Liang for sharing his software and experimental results with us. Finally, we thank the anonymous reviewers of JAIR and the editor, Lillian Lee, for their insightful comments which helped improve the final presentation of this paper. This work was funded by the NSF grant IIS–0712907X. Most of the experiments were run on the Mastodon Cluster, provided by NSF Grant EIA-0303609.

## Appendix A. Details of the meaning representation language

Table 10 shows brief explanations of the different events we detect with our simulated perception.

| Event | Description |
|---|---|
| Playmode | Signifies the current play mode as defined by the game |
| Ballstopped | The ball speed is below a minimum threshold |
| Turnover | The current possessor of the ball and the last possessor are on different teams |
| Kick | A player having possession of the ball in one time interval and not in the next |
| Pass | A player gains possession of the ball from a different player on the same team |
| BadPass | A pass in which the player gaining possession of the ball is on a different team |
| Defense | A transfer from one player to an opposing player in their penalty area |
| Steal | A player having possession of the ball in one time interval and another player on a different team having it in the next time interval |
| Block | Transfer from one player to the opposing goalie. |

Table 10: Description of the different events detected

Below we include the context-free grammar we developed for our meaning representation language. All derivations start at the root symbol *S.

```
*S -> playmode ( *PLAYMODE )
*S -> ballstopped
*S -> turnover ( *PLAYER , *PLAYER )
*S -> kick ( *PLAYER )
*S -> pass ( *PLAYER , *PLAYER )
*S -> badPass ( *PLAYER , *PLAYER )
*S -> defense ( *PLAYER , *PLAYER )
*S -> steal ( *PLAYER )
*S -> block ( *PLAYER )
```

```
*PLAYMODE -> kick_off_l
*PLAYMODE -> kick_off_r
*PLAYMODE -> kick_in_l
*PLAYMODE -> kick_in_r
*PLAYMODE -> play_on
*PLAYMODE -> offside_l
*PLAYMODE -> offside_r
*PLAYMODE -> free_kick_l
*PLAYMODE -> free_kick_r
*PLAYMODE -> corner_kick_l
*PLAYMODE -> corner_kick_r
*PLAYMODE -> goal_kick_l
*PLAYMODE -> goal_kick_r
*PLAYMODE -> goal_l
*PLAYMODE -> goal_r

*PLAYER -> pink1
*PLAYER -> pink2
*PLAYER -> pink3
*PLAYER -> pink4
*PLAYER -> pink5
*PLAYER -> pink6
*PLAYER -> pink7
*PLAYER -> pink8
*PLAYER -> pink9
*PLAYER -> pink10
*PLAYER -> pink11
*PLAYER -> purple1
*PLAYER -> purple2
*PLAYER -> purple3
*PLAYER -> purple4
*PLAYER -> purple5
*PLAYER -> purple6
*PLAYER -> purple7
*PLAYER -> purple8
*PLAYER -> purple9
*PLAYER -> purple10
*PLAYER -> purple11
```